\def\year{2022}\relax
\documentclass[letterpaper]{article} 
\usepackage{aaai22}  
\usepackage{times}  
\usepackage{helvet}  
\usepackage{courier}  
\usepackage[hyphens]{url}  
\usepackage{graphicx} 
\usepackage{natbib}  
\usepackage{caption} 
\DeclareCaptionStyle{ruled}{labelfont=normalfont,labelsep=colon,strut=off} 
\usepackage{algorithm}
\usepackage{algorithmic}

\usepackage{newfloat}
\usepackage{listings}

\usepackage{amssymb}
\usepackage{amsmath}
\usepackage{array}
\usepackage{booktabs}
\usepackage{cite}
\usepackage{lipsum}

\floatstyle{ruled}
\newfloat{listing}{tb}{lst}{}
\floatname{listing}{Listing}
\title{
Semantic Segmentation for Point Cloud Scenes \\ 
via Dilated Graph Feature Aggregation and Pyramid Decoders}
\author{
    Yongqiang Mao\textsuperscript{\rm 1,2,3,4}, 
    Xian Sun\textsuperscript{\rm 1,2,3,4}, 
    Kaiqiang Chen\textsuperscript{\rm 1,2}\thanks{Corresponding author.},\\
    Wenhui Diao\textsuperscript{\rm 1,2,3,4}, 
    Zonghao Guo\textsuperscript{\rm 3,4}, 
    Xiaonan Lu\textsuperscript{\rm 1,2,3,4}, 
    Kun Fu\textsuperscript{\rm 1,2,3,4}
}
\affiliations{
    \textsuperscript{\rm 1}Aerospace Information Research Institute, Chinese Academy of Sciences, Beijing 100190, China\\
    \textsuperscript{\rm 2}Key Laboratory of Network Information System Technology (NIST), \\
    Aerospace Information Research Institute, Chinese Academy of Sciences, Beijing 100190, China\\
    \textsuperscript{\rm 3}University of Chinese Academy of Sciences, Beijing 100190, China\\
    \textsuperscript{\rm 4}School of Electronic, Electrical and Communication Engineering, \\
    University of Chinese Academy of Sciences, Beijing 100190, China\\

    \{maoyongqiang19, guozonghao19, luxiaonan19\}@mails.ucas.ac.cn, \\
    \{sunxian, chenkq, diaowh\}@aircas.ac.cn, kunfuiecas@gmail.com
}

\usepackage{bibentry}

\begin{document}
\maketitle

\begin{abstract}
Semantic segmentation of point clouds generates comprehensive understanding of scenes through densely predicting the category for each point. Due to the unicity of receptive field, semantic segmentation of point clouds remains challenging for the expression of multi-receptive field features, which brings about the misclassification of instances with similar spatial structures. 
In this paper, we propose a graph convolutional network DGFA-Net rooted in dilated graph feature aggregation (DGFA), guided by multi-basis aggregation loss (MALoss) calculated through Pyramid Decoders. 
To configure multi-receptive field features, DGFA which takes the proposed dilated graph convolution (DGConv) as its basic building block, is designed to aggregate multi-scale feature representation by capturing dilated graphs with various receptive regions. 
By simultaneously considering penalizing the receptive field information with  point sets of different resolutions as calculation bases, we introduce Pyramid Decoders driven by MALoss for the diversity of receptive field bases. 
Combining these two aspects, DGFA-Net significantly improves the segmentation performance of instances with similar spatial structures. 
Experiments on S3DIS, ShapeNetPart and Toronto-3D show that DGFA-Net outperforms the baseline approach, achieving a new state-of-the-art segmentation performance.    
\end{abstract}

\section{Introduction}
\begin{figure}[t]
    \setlength{\abovecaptionskip}{0.cm}
    \setlength{\belowcaptionskip}{-0.cm}
    \centering
    \includegraphics[width=1.0\columnwidth]{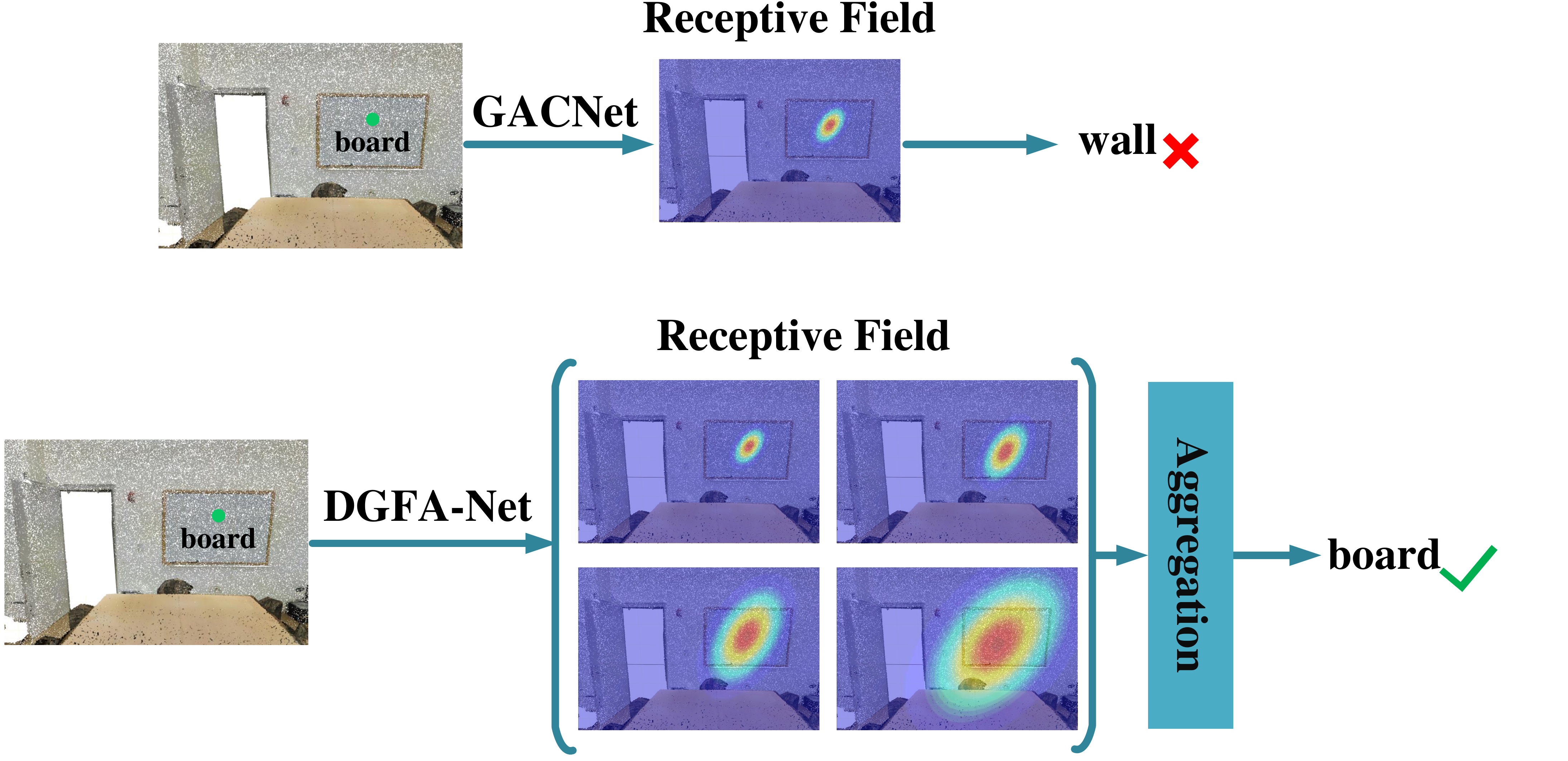} 
    \caption{Motivation: Sufficient multi-receptive field information is essential 
    for a thorough grasp of instances (such as board and wall) with similar spatial structures. 
    Here, we only illustrate the multi-level receptive fields around the green point "board" for visual clearness.
    Point "board" is misidentified as "wall" by GACNet on account of the unicity of receptive 
    fields, while with the aggregation of multi-receptive field information, it can be accurately 
    distinguished between wall and board by our DGFA-Net.}
\end{figure}

With the rapid development of 3D data acquisition techniques (such as LiDAR scanners and RGB-D cameras), a large amount of 3D data is applied to real life \cite{qi2018frustum}. Among all kinds of 3D data, point clouds play an essential role and draw a lot of attention. 
Although many previous point-based methods \cite{qi2017pointnet,qi2017pointnet++,mao2022beyond} have addressed the point cloud segmentation and achieved good results, the unicity of receptive field limits the expression of multi-receptive field features, leading to the misclassification of instances with similar spatial structures, Fig. 1(above). 

The universal point cloud segmentation networks \cite{qi2017pointnet++,wang2019dynamic} employ the fixed KNN graph to search for neighboring points, so that the receptive field obtained by using input point set as the calculation basis is limited to a single region and multi-scale local relationships cannot be captured, which cause the network to fail to learn relationships between points with similar spatial structures (such as adjacent planar objects: wall and board). 
It can be seen that the expression of multi-receptive field features is urgently needed. 
However, the expression of multi-receptive field features still has the following challenges:

\textbf{Unicity of the receptive field size.}
Most current methods \cite{thomas2019kpconv, wang2019graph} extract the point feature based on its pre-defined neighbors obtained by a fixed KNN graph. However, the difficulty in distinguishing similar regions caused by the determinism and unicity of the center point's receptive field has never been resolved. 
In the human visual system, contrast phenomenon refers to the phenomenon that when two stimuli act on the human eye at the same time, the presence of one stimulus enhances the other stimulus.
Since instances with similar spatial structures are spatially dependent, multi-scale receptive fields which capture multiple stimuli (instances) at the same time facilitate the distinction between these stimuli.
Inspired by this, we introduce a novel dilated graph convolution (DGConv) to obtain different scales of receptive fields, which captures the dilated graphs of different receptive regions through the proposed Sparse-KNN Search strategy according to the preset dilation rate. 
Further, to optimize the expression of multi-receptive field features, we propose a dilated graph feature aggregation (DGFA) module taking DGConv as the basic block for solving the misclassification of similar regions.

\textbf{Unicity of the receptive field calculation basis.}
The calculation basis of the existing 3D segmentation networks' \cite{qi2017pointnet++, thomas2019kpconv} receptive fields is the input point set, ignoring the pivotal role of point sets with different resolutions as the calculation bases of the receptive field. 
Inspired by the fact that human eyes can distinguish objects according to different references, we regard the point set of each resolution as a calculation basis for the receptive field and introduce Pyramid Decoders that decodes the feature into the corresponding resolution. 
Based on this, we introduce multi-basis aggregation loss (MALoss) which is calculated through the outputs of Pyramid Decoders.

To summarize, the contributions of this study include: 
First, we introduce dilated graph feature aggregation (DGFA) module taking the proposed DGConv as the basic block, which defines a systematic way to aggregate features with multi-scale receptive fields, Fig. 1(below). 
Second, we present Pyramid Decoders driven by multi-basis aggregation loss (MALoss), which achieves the effect of leveraging multi-level receptive field information with point sets of different resolutions as calculation bases. 
Combining these two aspects, we significantly improve the performance upon the baseline method, and achieve new state-of-the-art on commonly used benchmarks, especially for the instances with similar spatial structures.  
This suggests that \textbf{graph relationship between multiple stimuli (instances) in multi-scale receptive regions can significantly improve the segmentation performance of similar regions in point cloud scenes}. 
To the best of our knowledge, this is the first work to take the expression of multi-receptive field features into the distinguishment of instances with similar spatial structures in an explicit way. 

\section{Related Works}
\subsection{Point Clouds Feature Representation}
Taking into account the shortcomings of projection-based methods \cite{su2015multi} and voxel-based methods \cite{le2018pointgrid}, 
PointNet \cite{qi2017pointnet} that directly processes point clouds is proposed. 
PointNet independently learns the features of each point and then uses the symmetric function (maxpooling) to aggregate global features. 
On this basis, a large number of point-based methods have sprung up. 
Point-wise MLP methods \cite{qi2017pointnet++, jiang2018pointsift} use shared MLP as the basic unit of the network for feature extraction. 
Point Convolution Methods \cite{liu2019densepoint, atzmon2018point} aim to extract high-quality point set features and learn local relationships by designing efficient point convolution operators. 
However, the mentioned methods basically focus on the design of efficient point convolution operators, ignoring the geometric relationship between adjacent points and the unicity of the feature receptive field in the segmentation task itself. 



\subsection{Dilated Convolution on 3D Point Clouds}
In the image segmentation, dilated convolution \cite{holschneider1990real} is proposed to expand the receptive field of features. 
Similarly, Dilated convolution in GCNs \cite{li2019deepgcns} of point cloud segmentation is employed through selecting $k$ vertices formed by grabbing one point every $r$ (dilation rate) points from the KNN graph. 

Nevertheless, the vertices obtained by this method are too sparse and cannot express the spatial and geometric relationships of similar points well. 
To solve the sparseness of graphs, we propose a dilated graph convolution that employs a Sparse-KNN Search strategy for step sampling.

\subsection{Multi-scale Feature Expression}
The irregularity and unevenness of point clouds bring great challenges to the multi-scale expression of features. PointNet++ \cite{qi2017pointnet++} uses a hierarchical structure to extract the local features of the point set. 
From the perspective of multi-scale feature expression, multi-scale grouping and multi-resolution grouping are also proposed to overcome the problem of uneven density.
Later, Pointwise Pyramid Pooling module \cite{ye20183d} is first introduced to capture the coarse-to-fine local feature of multi-scale point sets.

Nonetheless, they still suffer from the limitation of small neighboring regions and cannot capture local-to-global features. 
Aiming to make full use of the advantages of dilated convolution to obtain multi-scale feature expression, our approach defines a more feasible way by employing DGConv. 
We apply a dilated graph feature aggregation module with DGConv as the basic block to obtain multi-scale receptive field feature representation. 
On this basis, we introduce Pyramid Decoders to fuse multi-level receptive field information with point sets of different resolutions as calculation bases.

\section{Proposed Approach}
\begin{figure*}[htb]
    \setlength{\abovecaptionskip}{0.cm}
    \setlength{\belowcaptionskip}{-0.cm}
    \centering
    \includegraphics[width=0.95\textwidth]{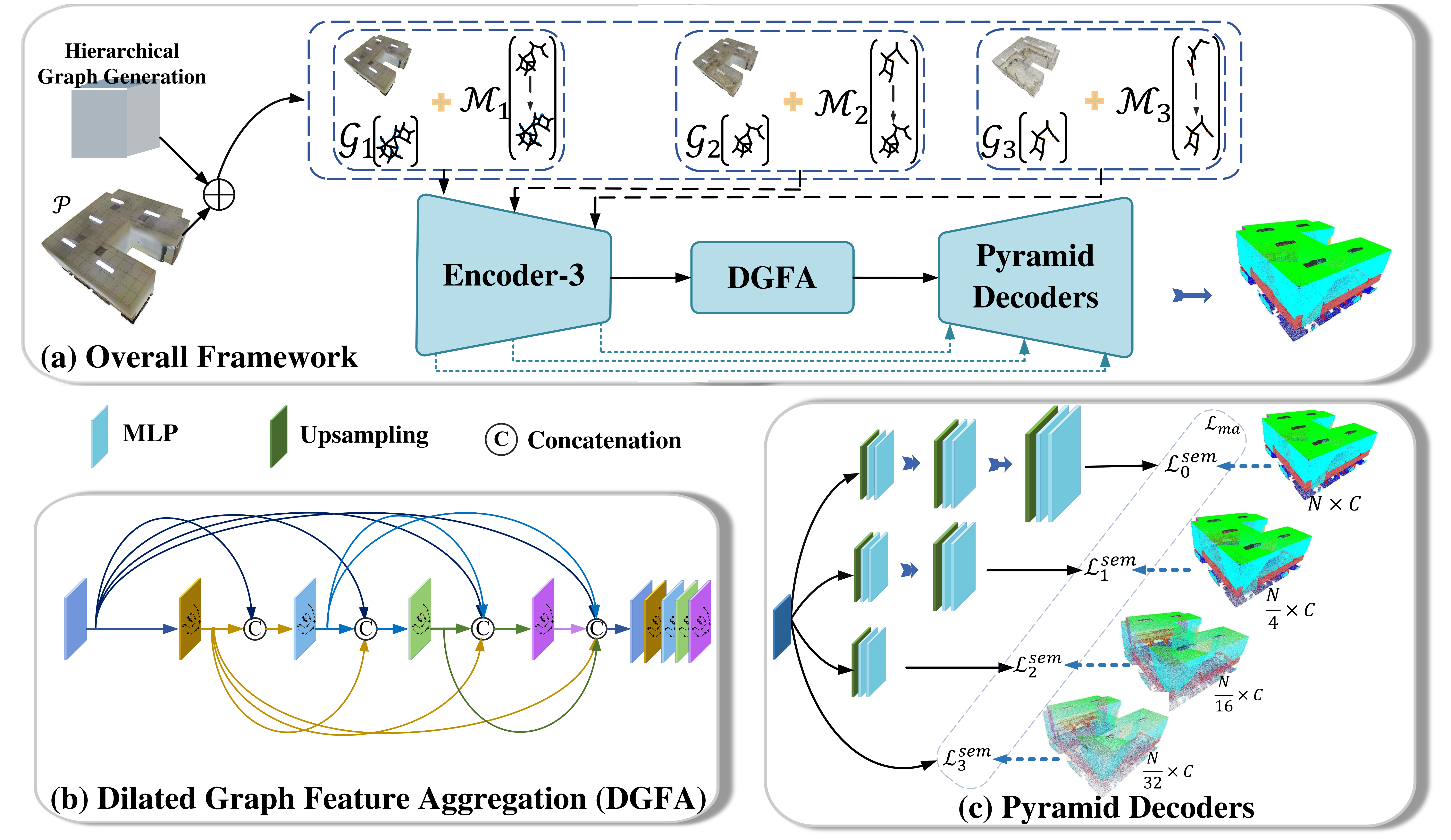} 
    \caption{Flowchart of the proposed DGFA-Net. 
    The framework (a) shows the two stages: hierarchical graph generation and Encoder-Decoder feature extraction and aggregation. 
    The outputs of hierarchical graph generation are point sub-graphs $\mathcal{G}=\{\mathcal{G}_{1}, \mathcal{G}_{2}, \mathcal{G}_{3}\}$ 
    and sampling mapping graphs $\mathcal{M}=\{\mathcal{M}_{1}, \mathcal{M}_{2}, \mathcal{M}_{3}\}$, which are each layer input of Encoder-3 
    (only 3 downsampling layers in encoder). 
    The Encoder-Decoder feature extraction and aggregation part consists of Encoder-3, the proposed DGFA module (b) and Pyramid Decoders (c) with MALoss $\mathcal{L}_{ma}$.}
\label{fig2}
\end{figure*}
\subsection{Overview}
The flowchart illustrated in Fig. 2(a) shows the whole architecture of DGFA-Net. 
Our DGFA-Net utilizes GACNet \cite{wang2019graph} with three downsampling layers as the baseline.
As a point cloud segmentation network, DGFA-Net is composed of two stages: hierarchical graph generation and Encoder-Decoder feature extraction and aggregation. 
The input of DGFA-Net is hierarchical graphs generated in advance, and features of a raw point set. 
In the first stage, the raw point set $\mathcal{P}=\{p_{i}\in \mathbb{R}^{3}, i=1,2,\cdots, N\}$ is sent to hierarchical graph generation 
to define point sub-graphs $\mathcal{G}=\{\mathcal{G}_{1}, \mathcal{G}_{2}, \mathcal{G}_{3}\}$ and sampling mapping graphs 
$\mathcal{M}=\{\mathcal{M}_{1}, \mathcal{M}_{2}, \mathcal{M}_{3}\}$, (above Encoder-3 in Fig. 2(a)). 
The second stage is responsible for the representation and aggregation of point features. 
The two critical modules are Dilated Graph Feature Aggregation (Fig. 2(b)) to solve the unicity of the receptive field size and Pyramid Decoders (Fig. 2(c)) to settle the unicity of the receptive field calculation basis. 

\subsection{Dilated Graph Feature Aggregation Module}
Analogous to the human visual system, contrast effect applied to point convolution can better enhance the central stimulus (instance) of two or more stimuli (instances).
Therefore, we propose a novel dilated graph convolution (DGConv) (Fig. 3) to guarantee features to obtain receptive fields with adjacent stimuli of different scales and add dilated graph feature aggregation (DGFA) with DGConv as the basic block to solve the unicity of the receptive field size.

\subsubsection{Dilated Graph Convolution}
Considering the graph $\mathcal{G}(\mathcal{V},\mathcal{E})$, $\mathcal{V}=\{v_{1},\cdots,v_{N}\}$ and $\mathcal{E}\subseteq |\mathcal{V}| \times |\mathcal{V}|$ 
represent the set of vertices and edges respectively. 
To get a larger receptive field, we propose a novel 
sparse $k$-nearest neighbor search (Sparse-KNN Search) strategy. 
Different from the ordinary $k$-nearest neighbor, our Sparse-KNN Search selects $\mathcal{K}_{s}$ nearest neighbors in the expansion region according to the dilation rate $r$, which can be formulated as follows:
\begin{equation}
\begin{aligned}
    \mathcal{K}_{s}=&\left \lfloor \frac{\mathcal{K}}{\Delta } \right \rfloor * (r -1 + \Delta) + \\
    &\left \lceil( \frac{\mathcal{K}}{\Delta} - \left \lfloor \frac{\mathcal{K}}{\Delta } \right \rfloor)*(r - 1 + \Delta)  \right \rceil
\end{aligned}
\end{equation}
where $ \left \lfloor \right \rfloor$ is the round down operation,  $ \left \lceil \right \rceil$ is the round up operation and $\Delta$ indicates the sampling step.
As claimed in Alg. 1, for each point set $\mathcal{P}=\{p_{i}\in \mathbb{R}^{3}, i=1,2, \cdots, N\}$, our Sparse-KNN Search strategy is to sample $\Delta$ points by skipping every $r$ points based on the selected $\mathcal{K}_{s}$ neighbors.

After that, we construct a novel dilated graph $\mathcal{G}(\mathcal{V},\mathcal{E}^{r})$. The new edges $\mathcal{E}^{r}$ of the vertex $v_{i}$ is written as:
\begin{equation}
    \begin{aligned}
    \mathcal{E}^{r}_{i} = &\{e_{i,r},\cdots, e_{i,r+\Delta-1}\}\\
    &\cup\{e_{i,2r+\Delta-1}, \cdots, e_{i,2r+2(\Delta-1)}\}\\
    &\cup \cdots \cup \{e_{i,mr+(m-1)(\Delta-1)}, \cdots, e_{i,\mathcal{K}_{s}}\}
    \end{aligned}
\end{equation}
where $e_{i,r}$ is the edge between the central vertex $v_{i}$ and its $r$-th nearest neighbor.
Denote $\mathcal{N}(i)=\{j:e_{i,j}\ \epsilon\ \mathcal{E}^{r}_{i}\}$ as the neighbor set of vertex $v_{i}$. 
Our DGConv is formulated as:
\begin{equation}
    \tilde{x_{i}} = \underset{j\in \mathcal{N}(i)}{max} \{\Theta (x_{i}, x_{j})\}
\end{equation}
where $\Theta$ represents 2D convolution and $max$ is feature aggregation function. 
$x_{i}$ indicates the feature of central vertex $v_{i}$ and $\{x_{j}, j\in \mathcal{N}(i)\}$ are features of neighbor vertex $v_{j}$.

\begin{figure}[t]
    \setlength{\abovecaptionskip}{0.cm}
    \setlength{\belowcaptionskip}{-0.cm}
    \centering
    \includegraphics[width=1.0\columnwidth]{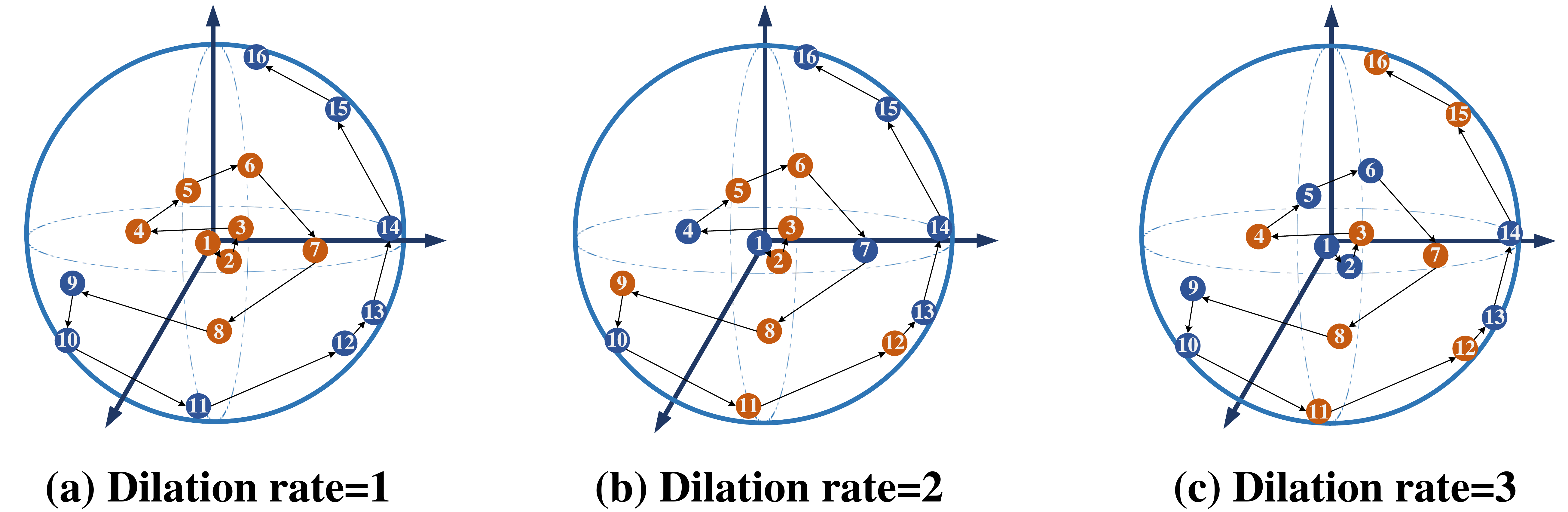} 
    \caption{The proposed dilated graph convolution (DGConv). 
    (a), (b) and (c) represent the results of the proposed Sparse-KNN Search strategy with a dilation rate of 1, 2 and 3, respectively. 
    The step sampling rate shown is 2. The number of target neighbors $\mathcal{K}$ (orange points) is 8.}
\label{fig3}
\end{figure}
\begin{algorithm}[tb]
\caption{Sparse-KNN Search Strategy}
\label{alg:algorithm}
\textbf{Input}: N points $\{\bf p_{1}, p_{2}, \cdots, p_{N}\}$\\
\textbf{Parameter}: number of neighbors $\mathcal{K}_{s}$ after expansion, number of target neighbors $\mathcal{K}$, sampling step $ \Delta $,
dilation rate $r$\\
\textbf{Output}: $\{\bf p_{1,\mathcal{K}},p_{2,\mathcal{K}},\cdots, p_{N,\mathcal{K}}\}$ N points with $\mathcal{K}$ neighbors
\begin{algorithmic}[1] 
\FOR{$\bf p_{m}\in \{\bf p_{1}, p_{2}, \cdots, p_{N} \}$}
\STATE $p_{m,\mathcal{K}}=\{\}$;
\STATE {\bf KNN-search: $\bf p_{m,\mathcal{\mathcal{K}}_{s}}\leftarrow p_{m}$};
\FOR{$i=1\ \textrm{to}\left \lceil \frac{\mathcal{K}_{s}}{r - 1 + \Delta}\right \rceil - 1$}
\STATE 
$a=(i-1) *(r + \Delta - 1) + r$\\
$b=i*(r + \Delta - 1)$\\
$p_{m,\mathcal{K}}=p_{m,\mathcal{K}} \cup p_{m,a:b}$;
\ENDFOR
\STATE
$i=\left \lceil \frac{\mathcal{K}_{s}}{r - 1 + \Delta}\right \rceil $\\
$a= ir + (i-1)*(\Delta -1)$\\
$b=\mathcal{K}_{s}$\\
$p_{m,\mathcal{K}}=p_{m,\mathcal{K}} \cup p_{m,a:b}$;
\ENDFOR
\STATE \textbf{return} $\forall p_{m,\mathcal{K}} \in \{p_{1,\mathcal{K}}, \cdots, 
p_{N,\mathcal{K}}\},p_{m,\mathcal{K}} \in \mathbb{R}^{1\times \mathcal{K}} $;
\end{algorithmic}
\end{algorithm}

\subsubsection{Dilated Graph Feature Aggregation}
With DGConv as the basic block, dilated graph feature aggregation employs the dense connection \cite{huang2017densely,yang2018denseaspp} mode to aggregate convolution outputs with different dilation rates. 
The structure of DGFA is illustrated in Fig. 2(b). 
DGConvs are organized in a cascade fashion, where the dilation rate of each layer increases layer by layer. 
DGConvs with small dilation rates are set in the shallower part, while DGConvs with larger dilation rates are set in the deeper part. 
The input of the following operation is the concatenated feature map of all the outputs of each shallower DGConv and the input feature map. 
The final output of DGFA is a feature map generated by multi-dilation rate DGConvs. 

Before concatenating the feature maps, each DGConv in DGFA can be formulated as follows:
\begin{equation}
    \mathcal{F}_{m}=\mathcal{G}^{m}_{r}\{\mathcal{T}\{\mathcal{F}_{0}, \mathcal{F}_{1}, \cdots , \mathcal{F}_{m-1}\}\}
\end{equation}
where $\mathcal{G}^{m}_{r}$ represents the $m$-th DGConv with dilation rate $r$, and $\mathcal{F}_{0}$ denotes the input feature map. 
$\mathcal{T}\{\cdot\}$ means the feature map formed by concatenating outputs from all previous DGConvs. 
Consequently, we write DGFA as:

\begin{equation}
    \mathcal{F}_{out}=\mathcal{M}\{\mathcal{R}\{\mathcal{F}_{0}, \mathcal{F}_{1}, \cdots , \mathcal{F}_{M}\}\}
\end{equation}
where $\mathcal{R}$ indicates the aggregation function concatenate operation and $\mathcal{M}$ indicates the multilayer perceptron. 

\subsection{Pyramid Decoders} 
The strategy to make the most of multi-level receptive fields in the imagery is to use different number of sampling layers, like Unet++ \cite{zhou2018unet++}. 
This approach is focusing on the receptive field information of different downsampling layers with the point set of original resolution as calculation basis.
Inspired by the fact that human eyes can distinguish objects according to different references, 
Pyramid Decoders driven by multi-basis aggregation loss, is introduced to settle the unicity of the receptive field calculation basis.

As illustrated in Fig. 4, the feature decoding part of DGFA-Net is rooted in Pyramid Decoders with different upsampling layers. 
Specifically, the output of DGFA is respectively up-sampled back to the resolution size of different layers for per-point prediction. 
We use $\mathcal{F}^{i,j}$ to represent the feature map, where $i$ indexes the downsampling layers and $j$ indexes different feature maps in the same layer. 
Thus, each $\mathcal{F}^{i,j}$ can be computed as the following formula:
\begin{equation}
    \mathcal{F}^{i,j}=\left\{\begin{matrix}
    GAC(\mathcal{F}^{i-1,j}),&i>0, j=0 \\
    \mathcal{M}(\mathcal{F}^{i,0}\ ||\ \mathcal{F}^{i+1,j+1}), &i<2,j>0\\
    \mathcal{M}(\mathcal{F}^{i,0}\ ||\ \mathcal{F}^{i+1,1}), & i=2,j>0 \\
    DGFA(\mathcal{F}_{i,j-1}),&i=3,j=1\\
    \end{matrix}\right.
\end{equation}
where $GAC$ represents the feature extractor, $\mathcal{M}$ indicates the multilayer perceptron and $||$ is the concatenate operation. 
Meanwhile, \{$\mathcal{F}^{3,1}$, $\mathcal{F}^{2,1}$, $\mathcal{F}^{1,1}$, $\mathcal{F}^{0,1}$\} are the outputs with different resolutions, respectively.

\begin{figure}[t]
    \setlength{\abovecaptionskip}{0.cm}
    \setlength{\belowcaptionskip}{-0.cm}
    \centering
    \includegraphics[width=1.0\columnwidth]{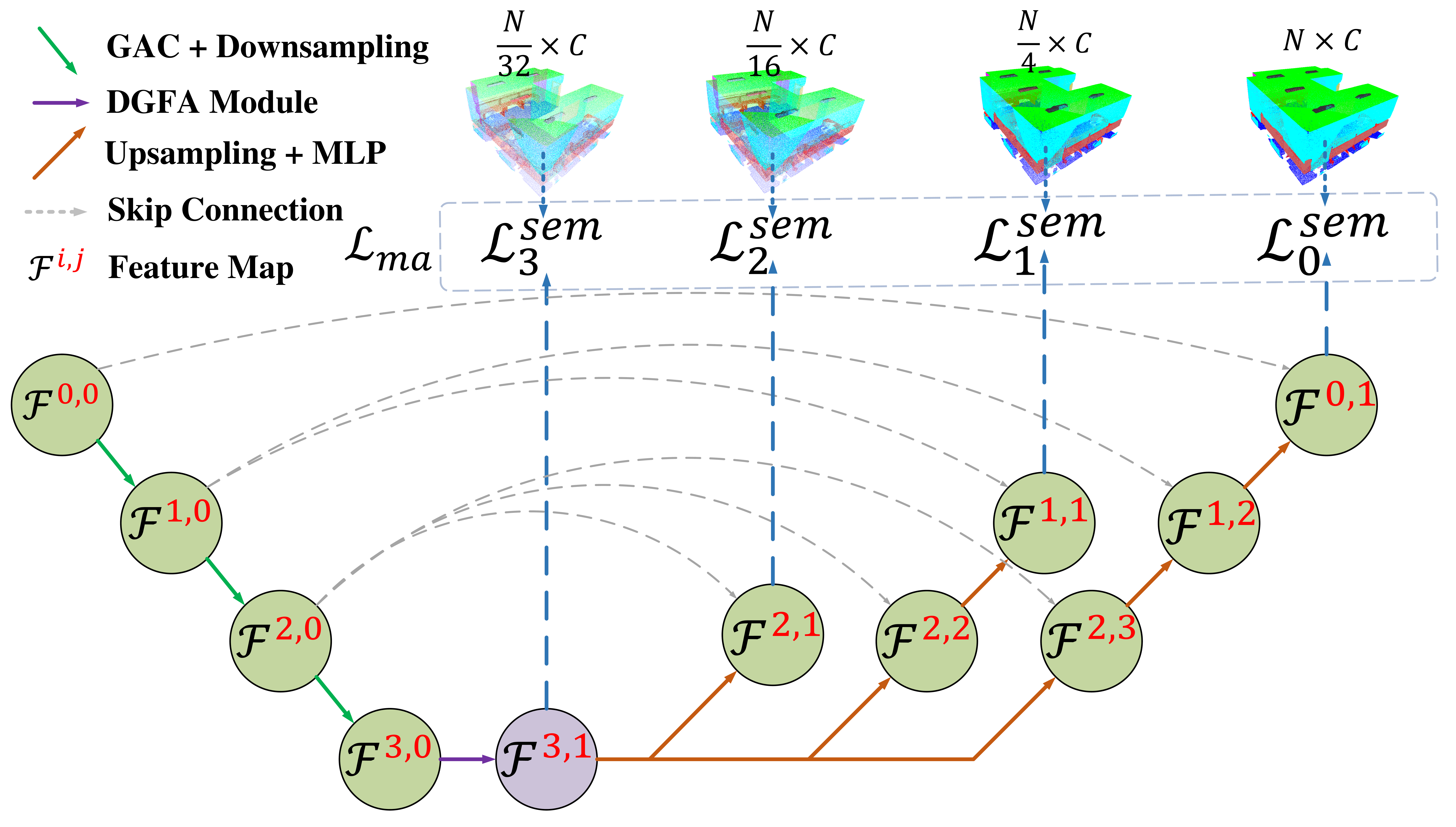} 
    \caption{Detailed diagram of DGFA-Net. 
    DGFA-Net consists of a shared encoder, a shared DGFA module and Pyramid Decoders. 
    Meanwhile, the encoder and Pyramid Decoders are connected through the skip pathways.}
\label{fig4}
\end{figure}

\begin{table*}[htb]
    \small            
    \setlength{\abovecaptionskip}{0pt}
    \caption{Semantic segmentation results (\%) on S3DIS Area-5. 
    Pre-proc. denotes the sampling method in data pre-processing. 
    Block and Grid represent the network uses block sampling method and grid sampling method, respectively.}
    \centering
            \begin{tabular}{m{7.0cm}|m{2.0cm}<{\centering}|m{2.0cm}<{\centering}m{2.0cm}<{\centering}m{2.0cm}<{\centering}}
            \toprule
                Method &Pre-proc.&mIoU &OA &mAcc \\  
            \midrule
                KPConv rigid \cite{thomas2019kpconv}        &Grid            & 65.4                     & -             & 70.9           \\ 
                KPConv deform \cite{thomas2019kpconv}       &\bf{Grid}       & \bf{67.1}                & -             & 72.8           \\ 
                
                PointNet \cite{qi2017pointnet}              &Block           & 41.1                     & -             & 49.0           \\ %
                SPG \cite{landrieu2018large}                &Block           & 58.0                     & 86.4          & 66.5           \\ 
                PointWeb \cite{zhao2019pointweb}            &Block           & 60.3                     & 87.0          & 66.6           \\ 
                HPEIN \cite{jiang2019hierarchical}          &Block           & 61.9                     & 87.2          & 68.3           \\ 
                \midrule
                baseline                                 &Block           & 61.8                     & 87.0          & 70.3       \\ 
                \bf{DGFA-Net (ours)}                                &\bf{Block}      & \bf{65.8}(4.0$\uparrow$) & \bf{88.2}(1.2$\uparrow$) & \bf{73.8}(3.5$\uparrow$) \\
            \bottomrule
    \end{tabular}
\end{table*}

\begin{table}[htb]
    \small
    \setlength{\abovecaptionskip}{0pt}
    \caption{Part segmentation results (\%) on ShapeNetPart.} 
    \centering
            \begin{tabular}{m{4.8cm}@{}@{}m{1.3cm}<{\centering}m{1.3cm}<{\centering}}
            \toprule
                Method &Class mIoU &Instance mIoU \\
            \midrule
                PointNet \cite{qi2017pointnet}             & 80.4      & 83.7\\ 
                PointNet++ \cite{qi2017pointnet++}           & 81.9      & 85.1\\
                PCNN \cite{wang2018deep}                & 81.8      & 85.1\\
                DGCNN \cite{wang2019dynamic}               & 82.3      & 85.1\\ 
                PointConv \cite{wu2019pointconv}            & 82.8      & 85.7\\ 
                InterpCNN \cite{mao2019interpolated}           & 84.0      & 86.3\\
                KPConv \cite{thomas2019kpconv}              & \bf{85.1} & \bf{86.4}\\
            \midrule  
                baseline          & 83.0      & 85.5       \\
                \bf{DGFA-Net (ours)}       & 83.8($0.8\uparrow$)& 86.3($0.8\uparrow$)  \\ 
            \bottomrule
    \end{tabular}
\end{table}

\subsubsection{Multi-basis Aggregation Loss}
Before feeding the input point set into the network, labels of different resolutions $\{\mathcal{A}_{0}\in \mathbb{R}^{N\times C}, \mathcal{A}_{1}\in \mathbb{R}^{N/4\times C}, \mathcal{A}_{2}\in \mathbb{R}^{N/16\times C}, \mathcal{A}_{3}\in \mathbb{R}^{N/32\times C}\}$ are sampled according to indexes of farthest point sampling. 
Using the outputs of Pyramid Decoders, fully connected layers predict confidence scores for all candidate semantic classes. 
Generally, we use cross-entropy loss $\mathcal{L}^{sem}$ as the semantic segmentation loss. 
For making full use of multi-level receptive fields, multi-basis aggregation loss $\mathcal{L}_{ma}$ is calculated by each output $\{\mathcal{\hat{S}}_{0}, \mathcal{\hat{S}}_{1},\mathcal{\hat{S}}_{2},\mathcal{\hat{S}}_{3}\}$ of Pyramid Decoders and the label with corresponding resolution. 
Hence, $\mathcal{L}_{ma}$ is formulated as follows:
\begin{equation}
    \begin{aligned}
        &\mathcal{L}_{ma}=\sum_{i=0}^{3}\lambda_{i}\mathcal{L}^{sem}_{i} 
        =\sum_{i=0}^{3}\lambda_{i} \sum_{j=1}^{N_{i}}\sum_{c=1}^{C}\mathcal{A}^{jc}_{i}log \mathcal{\hat{S}}^{jc}_{i} 
    \end{aligned}
\end{equation}
where $\lambda_{i}$ is the weight hyperparameter for the loss of each decoder, $N_{i}$ is 
the number of point sets with different resolutions and $C$ indicates the number of categories.

\section{Experiments}
In this section, we conduct experiments and evaluate our DGFA-Net on three point cloud segmentation benchmarks, 
including Stanford Large-Scale 3D Indoor Spaces (S3DIS) \cite{armeni20163d}, ShapeNetPart \cite{yi2016scalable}, and Toronto-3D \cite{tan2020toronto} dataset. We utilize GACNet with three downsampling layers as the baseline. 
To quantitatively evaluate the performance of our DGFA-Net, three metrics are mainly used, including mean Intersection-over-Union (mIoU), overall accuracy (OA) and average class accuracy (mAcc). 
Our model is trained on a single GeForce RTX 3090 GPU with batch size 16 for each dataset. 
Adam optimizer is employed to minimize the overall loss in Eq. 7.

\subsection{Datasets}
\subsubsection{S3DIS}
The S3DIS dataset collects 3D RGB point clouds including 271 rooms. 
For a principled evaluation, we follow PointNet++ \cite{qi2017pointnet++} to choose Area-5 as our testing set. 
To prepare our training data, we first split the dataset room by room and then sample them into 1.0m $\times$ 1.0m blocks. 
Subsequently, the hyperparameters $\{\lambda_{0}, \lambda_{1}, \lambda_{2}, \lambda_{3}\}$ for $\mathcal{L}_{ma}$ are set to $\{1.0, 1.5, 2.0, 2.5\}$.
\begin{table}[htb]
    \small
    \setlength{\abovecaptionskip}{0pt}
    \caption{Semantic segmentation results (\%) on Toronto-3D.} 
    \centering
            \begin{tabular}{m{4.2cm}@{}@{}m{1.5cm}<{\centering}m{1.6cm}<{\centering}}
            \toprule
                Method &OA &mIoU \\
            \midrule
            PointNet++ \cite{qi2017pointnet++}         &91.21           & 56.55   \\ 
            PointNet++ (MSG)    &90.58           & 53.12   \\ 
            DGCNN \cite{wang2019dynamic}              &89.00           & 49.60 \\ 
            KPConv \cite{thomas2019kpconv}             &91.71           & 60.30  \\ 
            MS-PCNN \cite{ma2019multi}            &91.53           & 58.01  \\ 
            TGNet \cite{li2019tgnet}            &91.64           & 58.34 \\ 
            MS-TGNet \cite{tan2020toronto}           &91.69           & 60.96   \\ 
            \midrule  
            baseline              &88.94           & 57.06  \\
            \bf{DGFA-Net (ours)}          &\bf{94.78}(5.84$\uparrow$)  &  \bf{64.25}(7.19$\uparrow$)\\ 
        \bottomrule
\end{tabular}
\end{table}

\subsubsection{ShapeNetPart}
There are a total of 16,881 models in the ShapeNetPart dataset. 
These models are divided into 16 categories and 50 parts are marked. 
The goal of the segmentation task is to assign a part category to each point and give a shape category for each input point set.
Generally, 2048 points of each shape are sampled and fed into the network.
Moreover, we set the weight hyperparameters $\{\lambda_{0}, \lambda_{1}, \lambda_{2}\}$ for MALoss $\mathcal{L}_{ma}$ to $\{1.0, 0.5, 0.5\}$. 

\subsubsection{Toronto-3D}
The Toronto-3D dataset is collected from a region about 1 km on Avenue Road, Toronto and divided into 4 areas. 
Following \cite{tan2020toronto}, L002 is regarded as the testing set.
Further, the weight hyperparameters $\{\lambda_{0}, \lambda_{1}, \lambda_{2}, \lambda_{3}\}$ for $\mathcal{L}_{ma}$ are set to $\{1.0, 0.1, 0.1, 0.1\}$.

\subsection{Performance}
\subsubsection{Indoor Scene Segmentation}
As is depicted in Table 1, the proposed DGFA-Net is compared with other state-of-the-art methods on S3DIS. 
On the generalizability evaluation on S3DIS Area-5, our DGFA-Net outperforms PointNet by 24.7\% mIoU and 24.8\% mAcc. 
Further, DGFA-Net notably lifts the performance of baseline 
on mIoU (\textbf{4.0}\%$\uparrow $), OA (\textbf{1.2}\%$\uparrow $) and mAcc (\textbf{3.5}\%$\uparrow$), which are large margins. 
As a segmentation network with only 3 downsampling layers, our DGFA-Net reaches the best in all metrics among all methods which use block sampling to pre-process data. 
The reason lies in that DGFA-Net can capture local-to-global features of all instances, especially for instances with similar spatial positions and structures. 

\textbf{Visualization Analysis.} 
We visualize feature maps of each channel after the last upsampling layer, and the feature map after channel averaging for the analysis of planar instances, Fig. 7.
It is worth noting that the planar instances (ceiling, floor, wall and board) are accurately focused on. 
Especially, the two adjacent instances (wall and board) with similar spatial structures are well distinguished. 

As shown in Fig. 5, we present visualizations of the baseline and our DGFA-Net.
Our DGFA-Net achieves significantly better performance than the baseline in various scenes or areas.
The visualization results show the excellent performance of our method on various planar objects that are difficult to classify, 
including walls and boards.

\begin{figure}[t]
    \centering
    \includegraphics[width=1.0\columnwidth]{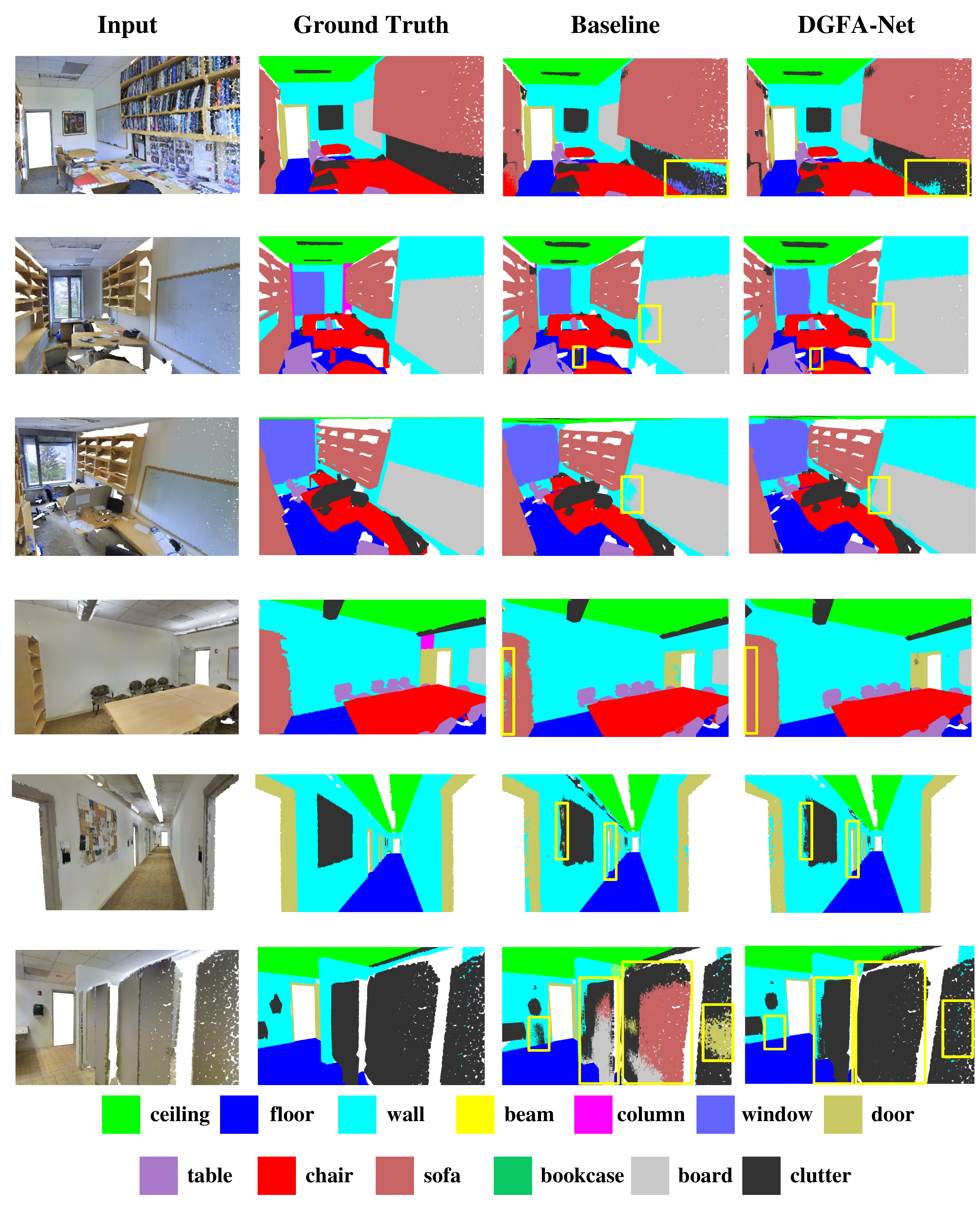} 
    \caption{Visualizations of S3DIS Area-5. 
    The first column shows original point cloud scene inputs, the second column shows the ground truth labels, 
    the third column shows the visualizations of the baseline segmentation results, 
    and the last column shows the scene visualizations of our DGFA-Net results. 
    From the comparison (shown in the yellow boxes) with the baseline, our DGFA-Net distinguishes similar planar instances better.}
    \label{fig6}
\end{figure}
\begin{figure}[t]
    \centering
    \includegraphics[width=1.0\columnwidth]{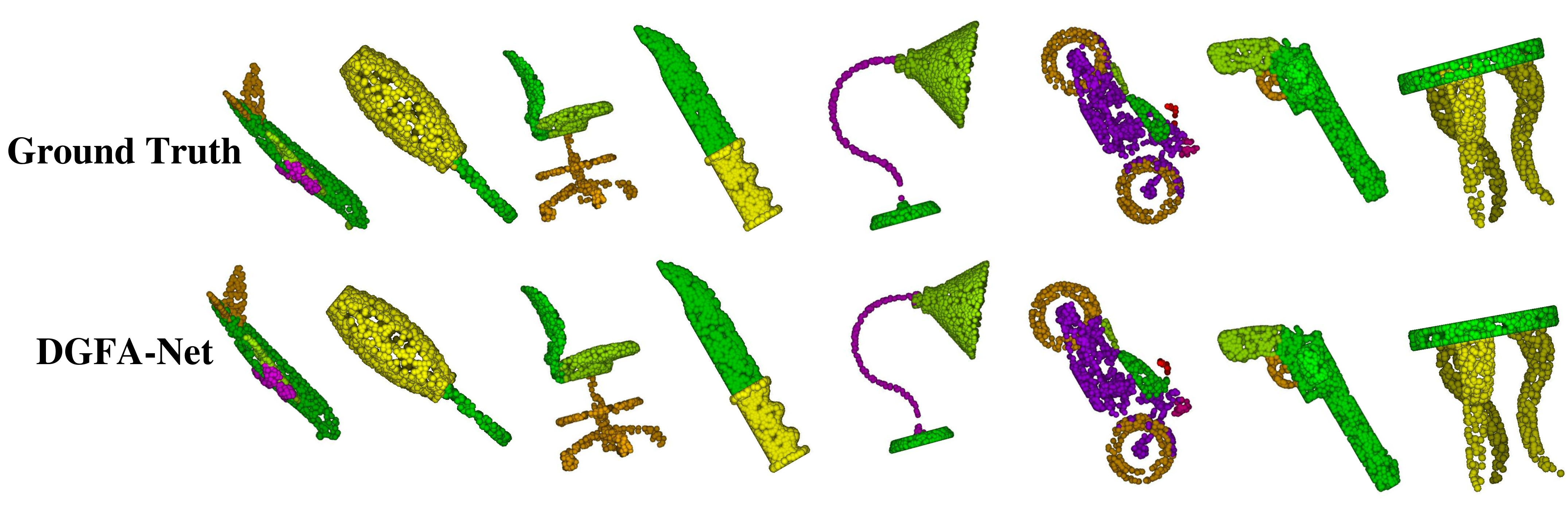} 
    \caption{Visualizations of ShapeNetPart. 
    The first row shows the ground truths and the second row shows the part visualizations of our DGFA-Net results.}
    \label{fig4}
\end{figure}
\subsubsection{Shape Part Segmentation}
In Table 2, we report the comparison results of our DGFA-Net and other state-of-theart part segmentation networks on ShapeNetPart. DGFANet achieves 83.8\% class mIoU and 86.3\% instance mIoU, reaching an improvement on the basis of the baseline. 
In general, the performance of our DGFA-Net is on par with or better than other competitive methods on both instance and class mIoU. 
It is noteworthy that the reason why we can get a superior instance mIoU is that our DGFA-Net can obtain receptive regions of different scales for a single object.
Besides, the reason for the low class mIoU is that ShapeNetPart is composed of independent 3D models without similar spatial structures. 
Additionally, our DGFA-Net surpasses or approaches other methods with much fewer parameters (analyzed in Ablation Study).

\textbf{Visualization Analysis.} 
In Fig. 6, visualizations of our DGFA-Net’s outputs and ground truths on ShapeNetPart are given. 
It is worth noting that our segmentation results are roughly consistent with the ground truths. 
The reason is that our network is analogous to the human visual system, covering adjacent stimuli (instances) in the receptive field of the central point at the same time.

\subsubsection{Outdoor Scene Segmentation}
As shown in Table 3, we list the mean IoU and overall accuracy of our DGFA-Net compared to other state-of-the-art methods on Toronto-3D. 
Although the encoder of our method has only 3 downsampling layers, DGFA-Net has improved OA from 88.94\% to 94.78\% and mIoU from 57.06\% 
to 64.25\%, which are great improvements. 
It's worth noting that our DGFA-Net outperforms all other methods on both indicators.

\subsection{Ablation Study}
To prove the effectiveness of our DGFA-Net, ablation studies are conducted on S3DIS \cite{armeni20163d}. 

\begin{figure}[t]
    \setlength{\abovecaptionskip}{0.cm}
    \setlength{\belowcaptionskip}{-0.cm}
    \centering
    \includegraphics[width=0.8\columnwidth]{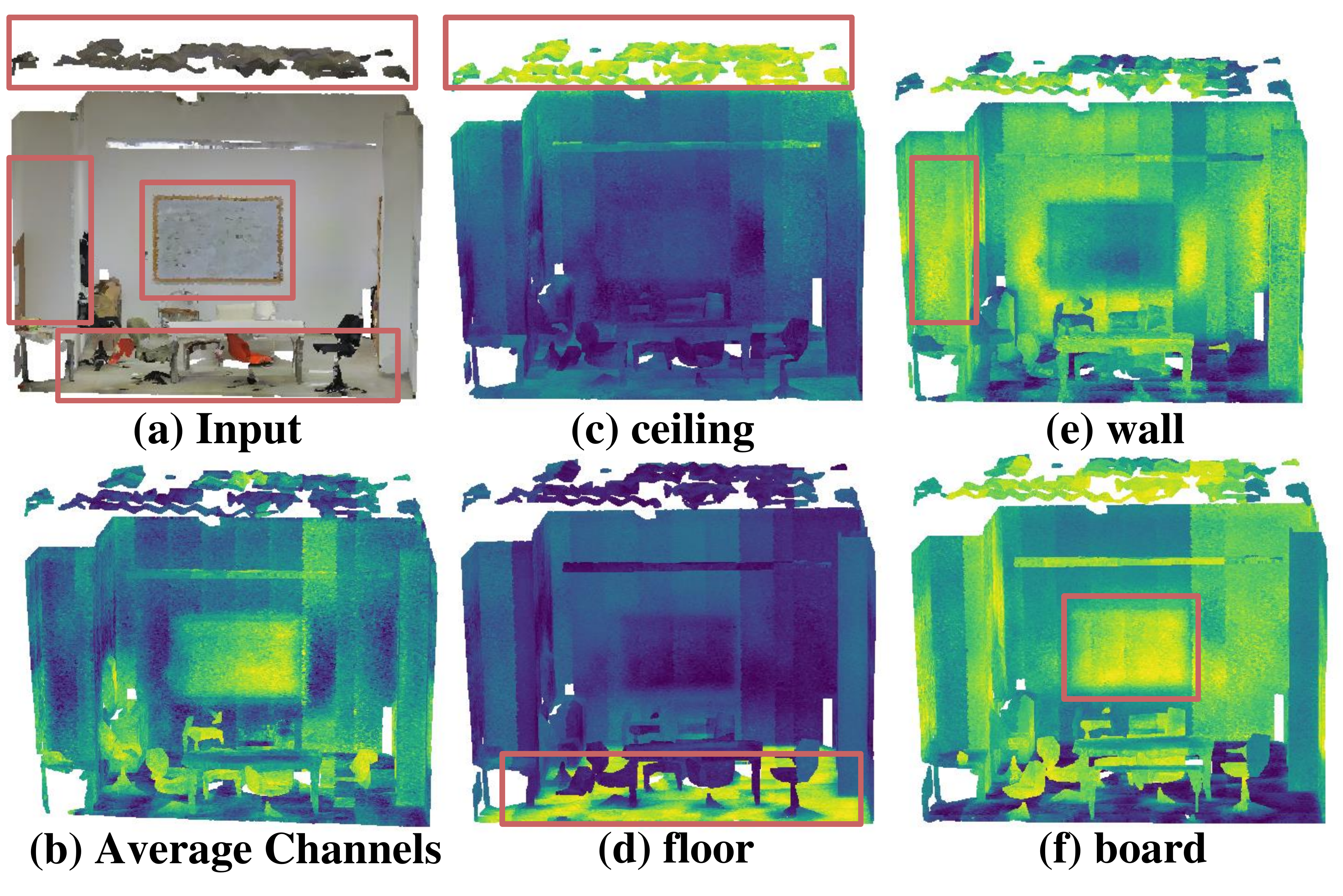} 
    \caption{The visualization of feature maps after the last upsampling layer. 
    The highlighted parts (red boxes) indicate that the relevant instances have received more attention. 
    By aggregating multi-receptive field features, the adjacent instances with similar spatial structures are well distinguished. }
    \label{fig5}
\end{figure}

\subsubsection{MALoss} 
In Table 4, with MALoss, DGFA-Net improves the performance by 0.8\% (62.6\% vs. 61.8\%), which validates that MALoss makes full use of 
the multi-level receptive field information by penalizing the outputs of Pyramid Decoders. 
Thus, minimizing MALoss drives better multi-scale feature expression. 
\begin{figure}[t]
    \centering
    \includegraphics[width=1.0\columnwidth]{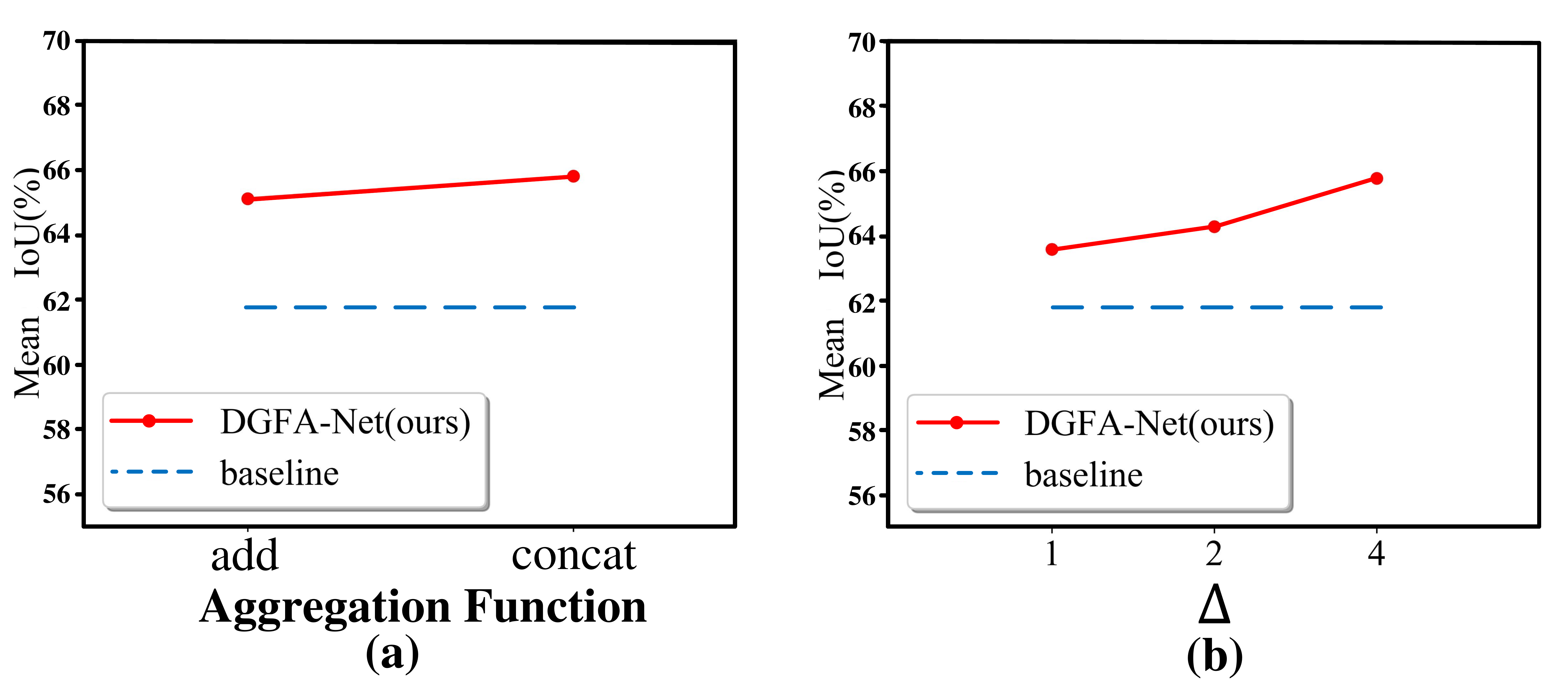} 
    \caption{Evaluation of hyperparameters and modules. 
    (a) Aggregation function. (b) Sampling step $\Delta$.}
    \label{fig4}
\end{figure}

\begin{table}[htb]
    \small  
    \setlength{\abovecaptionskip}{0pt}
    \caption{Ablation studies of the proposed modules. 
    "MALoss" denotes multi-basis aggregation loss, 
    "DGFA w/o (or w/) dense" denotes dilated graph feature aggregation without (or with) dense connection.}
    \centering
            \begin{tabular}{m{1.3cm}<{\centering} m{2.2cm}<{\centering} m{2.2cm}<{\centering}|
                m{1.1cm}<{\centering} }
            \toprule
            MALoss &DGFA w/o dense &DGFA w/ dense &mIOU\\
            \midrule
                                         &             &           &  61.8\\
                     $\surd  $            &             &           &  62.6  \\  
                    $\surd  $            &$\surd  $    &           &  63.8  \\  
                    $\surd  $            &$\surd  $    & $\surd  $ & \bf{65.8}\\
            \bottomrule
    \end{tabular}
\end{table}
\subsubsection{DGFA}
From Table 4, DGFA without dense connection improves the performance by 1.2\% (63.8\% vs. 62.6\%), which validates the feature extraction performance 
of multi-scale receptive fields is better than that of single receptive field. 
By adding the dense connection, DGFA with dense connection further improves the segmentation performance by 2.0\% (65.8\% vs. 63.8\%), 
which validates that the dense connection brings denser feature pyramid and larger receptive field. 
In total, DGFA improves the performance by 3.2\% (65.8\% vs. 62.6\%), which is a significant margin in semantic segmentation. 
This clearly demonstrates the superiority of the proposed DGFA over previous single receptive field methods. 
Besides, DGFA enhances the propagation of features and realizes the reuse of features. 

\subsubsection{DGConv.}
In the construction of DGConv dilated graphs, different step $\Delta$ values determine the different sparsity of dilated graphs. 
Through experiments, we prove that when the step $\Delta$ value is 4, the sparsity of the constructed dilated graphs can best meet the 
need of the network and our DGFA-Net reaches the best mIoU (65.8\%) (Fig. 8(b)).

\subsubsection{Aggregation Function.}
As shown in Fig. 8(a), we experimentally validate that using concatenate operation as the aggregation function ($\mathcal{R}$ of Eq. 5) of 
all DGConv outputs can achieve the best performance. 

\subsubsection{Dilation Combination} 
In Table 5, ablation study is carried out to determine the dilation combination of DGFA. The best performance occurs at dilation combination \{1, 2, 4, 8\}.
The reason lies in that the combination of more receptive regions enables the network to extract features of more scales.

\begin{figure}[t]
    \centering
    \includegraphics[width=0.8\columnwidth]{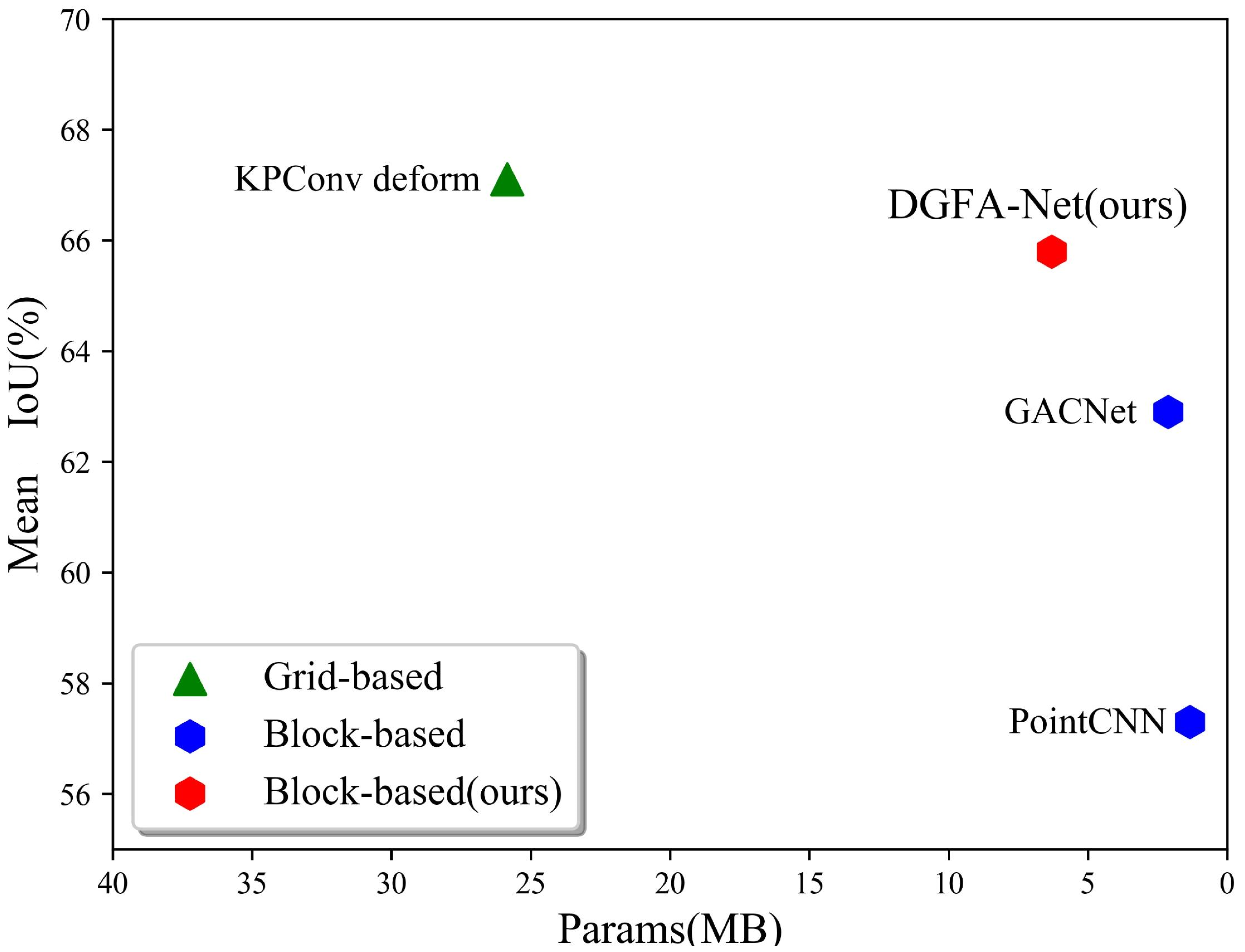} 
    \caption{The mIoU and parameters of different point segmentation methods. 
    Hexagons represent the methods using block sampling. 
    The red hexagon is our DGFA-Net, while blue hexagons are the existing methods GACNet and PointCNN. 
    The green triangle is the method KPConv which uses grid sampling.}
    \label{fig6}
\end{figure}

\begin{table}[htb]
    \small   
    \caption{Ablation study of dilation combination. "Dilation Rates" denotes the combinations of dilation rates.}
    \centering
    \begin{tabular}{m{3.0cm}<{\centering}|m{1.5cm}<{\centering}}
    \toprule
    Dilation Rates &mIOU \\  
    \midrule
    \{1,2\}                                                          & 63.7\\ 
    \{2,4\}                                                          & 64.7 \\ 
    \{1,4\}                                                          & 65.2 \\ 
    \{1,2,4,8\}                                                      & \bf{65.8} \\     
    \bottomrule
    \end{tabular}
\end{table}
\subsubsection{Network Complexity} 
DGFA-Net is a shallow segmentation network with 3 downsampling operations. 
As shown in Fig. 9, the existing segmentation methods introduce a lot of parameters. 
The extra parameters greatly limit their scope of application and practicality. In contrast, our DGFA-Net achieves the best performance in block sampling methods
while having fewer parameters.

\section{Conclusion}
We propose DGFA-Net, which is an elegant and effective network to extract multi-receptive field features for the distinguishment of instances with similar spatial structures. 
With DGFA, DGFA-Net implements the aggregation of multi-receptive field features in accordance with dilated graphs obtained from DGConv.
By optimizing MALoss, DGFA-Net realizes the optimized expression of the receptive field information with point set of each resolution as calculation basis.
Extensive experiments on commonly used benchmarks validate DGFA-Net's superior performance, in striking contrast with state-of-the-art 4-downsampling-layers segmentation networks. 
DGFA-Net provides a fresh insight for capturing multi-receptive field features.


\section{Acknowledgments}
This work is supported in part by National Natural Science Foundation of China (NSFC) under Grant 61725105. 

\bibliography{DGFA-Net}

\bigskip

\clearpage
\section{Supplementary Material}

\section{A. Overview}
The supplementary materials are arranged as follows:
First, we introduce the network details of our DGFA-Net.
Then, 
the details of segmentation results and visualization analyses on each dataset are given.

\section{B. Network Details}
From Fig. 2 in the body of our paper, our DGFA-Net consists of Pyramid Decoders while training. 
During inference, the decoder with three upsampling layers is retained, and other decoders that only play a role in training are discarded. 
Thus, the architecture of DGFA-Net is shown in Fig. 11. 
The Encoder-3 with 3 downsampling layers and the retained decoder with 3 upsampling layers complete the encoding and decoding of features, respectively. 
During the encoding of the features, the point cloud resolution of each layer is 
$N\rightarrow \frac{N}{4} \rightarrow \frac{N}{16} \rightarrow \frac{N}{32}$. 
For DGFA, the input and output feature dimensions are 1024 and 512 respectively. Moreover, the dilation combination of DGFA is $\{1,2,4,8\}$. Further, 
the sampling step of DGConv is set to 4. During the decoding of the features, point features are sequentially upsampled as 
$128\rightarrow 256 \rightarrow 1024 \rightarrow 4096$. 
\begin{figure}[t]
    \centering
    \includegraphics[width=1.0\columnwidth]{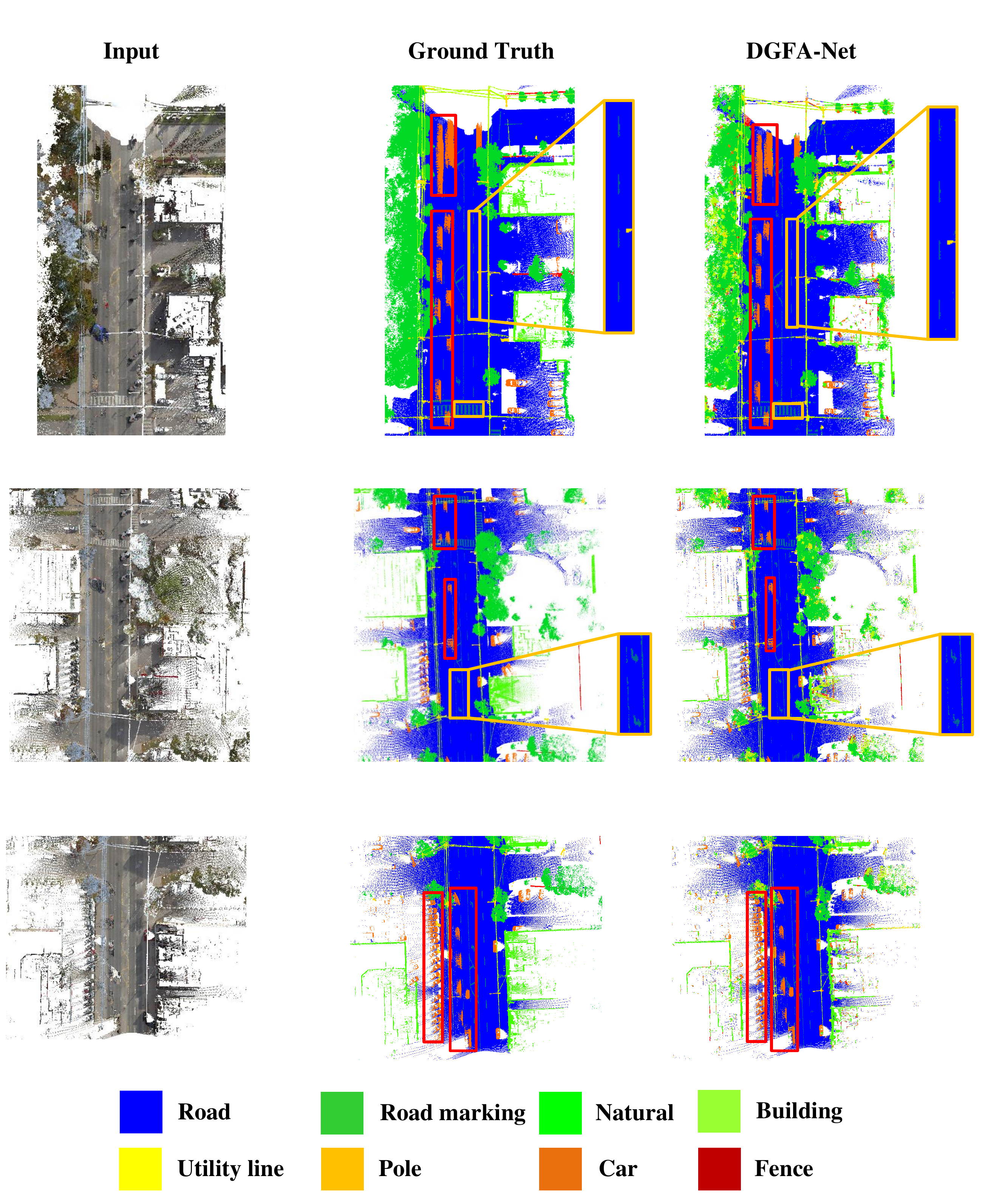} 
    \caption{Visualizations of Toronto-3D L002. 
    The first column shows original point cloud scene inputs, the second column shows the ground truth labels, 
    and the last column shows the scene visualizations of our DGFA-Net results. 
    From the comparison (shown in the yellow and red boxes) with the ground truths, 
    our DGFA-Net can distinguish instances with similar spatial structures better, 
    especially for the cars (red boxes) and road markings (yellow boxes).}
    \label{fig7}
\end{figure}
\begin{figure*}[htb]
    \centering
    \includegraphics[scale=0.15]{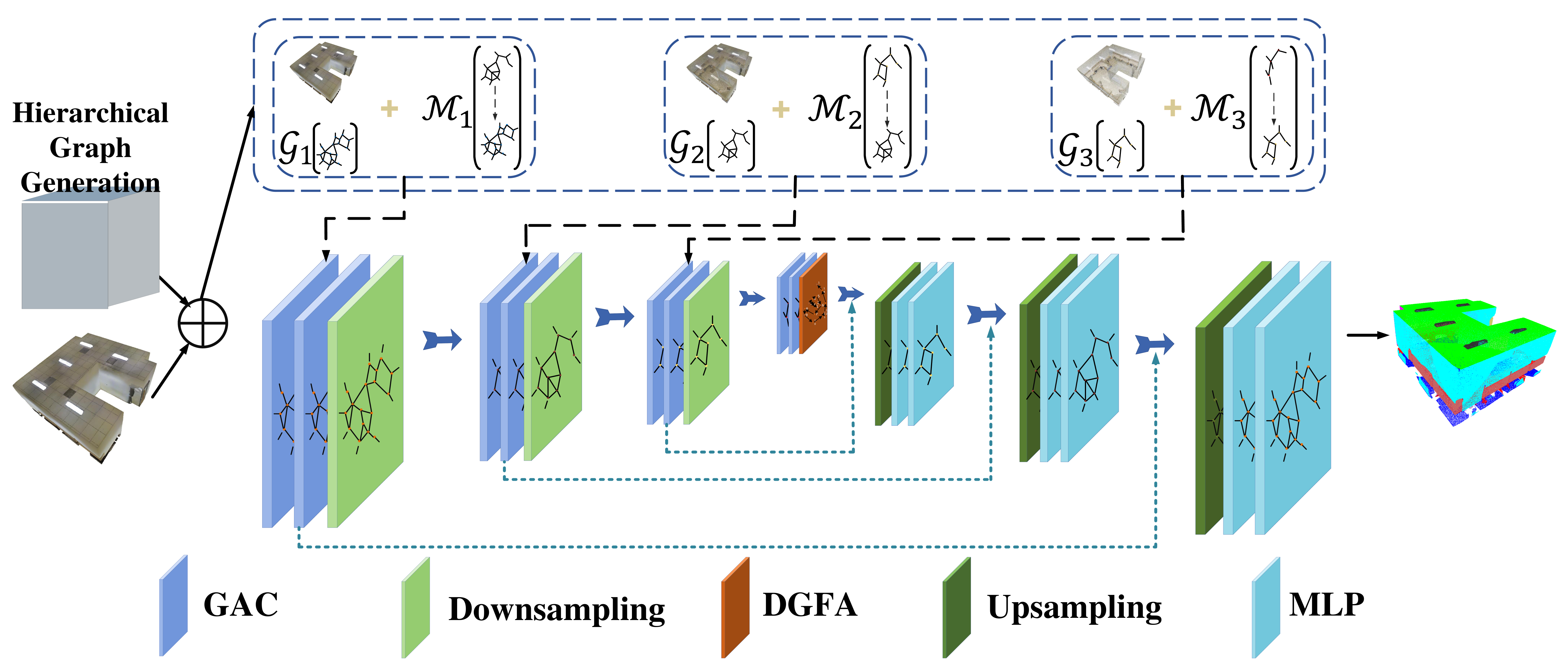}
    \caption{Flowchart of the proposed DGFA-Net during inference. 
    During a forward pass, the decoder with three upsampling layers is retained, and other decoders are discarded. 
    The features are extracted by consecutive operations (represented by cuboid colors). `GAC' represents Graph Attention Convolution.}
\label{fig3}
\end{figure*}
\section{C. More Segmentation Results}

\subsection{C.1. S3DIS}
\subsubsection{Segmentation Results}
As is depicted in Table 6, the segmentation results of each category on S3DIS Area-5 are given. 
Particularly, for the ceiling, floor, window, wall and board which are similar planar objects and difficult to classify, it achieves the best results. 
The reason lies in that the planar objects are similar in spatial structures and our DGFA-Net with multi-scale receptive fields can be more adaptive to such different planar objects.
Specifically, our DGFA-Net enables the feature receptive field to capture a larger adjacent region. 
Then, as long as the adjacent region contains a variety of similar planar objects at the same time, our DGFA-Net can better extract different features.

\subsection{C.2. ShapeNetPart}
\subsubsection{Segmentation Results}
Table 7 reports the results on ShapeNetPart, where our method achieves comparable or better performance than other part segmentation methods
on both instance and class mIoU. 
Especially for cap, chair, knife and lamp, our DGFA-Net achieves the highest classification performance among all other methods. 
Compared with the baseline, our method has comparable and higher performance in all categories. 
This shows the effectiveness of our proposed dilated graph feature aggregation (DGFA) module and multi-basis aggregation loss (MALoss).

\subsection{C.3. Toronto-3D}
\subsubsection{Segmentation Results}
Table 8 enumerates the mIoU of each category on Toronto-3D for outdoor roadway-level segmentation task. 
Especially for road marking and car that are difficult to classify, our DGFA-Net achieves the best performance. 
Notably, both road marking and car have similar spatial coordinate information as road.
Specifically, roads and road markings are similar in spatial position for the large outdoor point cloud scene.
Moreover, the spatial positions of cars and roads tend to be consistent compared to the overall scene scale. 
Since our DGFA-Net has multi-scale receptive fields and well retains the spatial scale information of the point set, 
it can better classify objects with similar spatial structures in larger scale point cloud scenes.

\subsubsection{Visualization Analysis} 
As illustrated in Fig. 10, we show the semantic segmentation performance on the large-scale point clouds of outdoor scenes. 
We compare our results against the ground truths on the testing set of Toronto-3D dataset. 
The red and yellow rectangles represent the visualization results of category car and road marking, respectively. 
Obviously, our DGFA-Net is well distinguished for road, car and road marking, which verifies the higher segmentation performance of our method for similar and adjacent planar instances.

\begin{table*}[htb]
    \footnotesize
    \caption{Detailed semantic segmentation results (\%) on S3DIS Area-5. 
    Pre. denotes the sampling method in data preprocessing. 
    Block and Grid represent the network uses block sampling method and grid sampling method, respectively.}
    \centering
    \begin{tabular}{m{2.2cm}@{}|@{}m{1.0cm}<{\centering}@{}|@{}m{0.9cm}<{\centering}@{}|@{}m{1.0cm}<{\centering}@{}|
        @{}m{0.95cm}<{\centering}@{}@{}m{0.95cm}<{\centering}@{}@{}m{0.95cm}<{\centering}@{}
        @{}m{0.95cm}<{\centering}@{}@{}m{0.95cm}<{\centering}@{}@{}m{0.95cm}<{\centering}@{}@{}m{0.95cm}<{\centering}@{}
        @{}m{0.95cm}<{\centering}@{}@{}m{0.95cm}<{\centering}@{}@{}m{0.95cm}<{\centering}@{}@{}m{0.95cm}<{\centering}@{}
        @{}m{0.95cm}<{\centering}@{}@{}m{0.95cm}<{\centering}@{}}
            \toprule
                Method &Pre.&mIoU &mAcc&ceil. &floor &wall  &beam  &col.  &wind.  &door  &chair  &table  &book  &sofa  &board  &clut.\\
            \midrule
                KPConv rigid        &Grid            & 65.4       & 70.9    & 92.6      &97.3       &81.4      &0.0           &16.5     &54.5        &\bf{69.5}          &90.1          &80.2          &74.6         &66.4         &63.7         &58.1\\ 
                KPConv deform       &\bf{Grid}       & \bf{67.1}  & 72.8    & 92.8      &97.3       &82.4      &0.0           &\bf{23.9}     &58.0        &69.0          &\bf{91.0}     &\bf{81.5}     &\bf{75.3}    &\bf{75.4}    &66.7         &\bf{58.9}\\ 
                PointNet            &Block           & 41.1       & 49.0    & 88.80     &97.33      &69.80     &\bf{0.05}     &3.92     &46.26       &10.76         &52.61         &58.93         &40.28        &5.85         &26.38        &33.22 \\ 
                SegCloud            &Block           & 48.9       & 57.4    & 92.31     &98.24      &79.41     &0.00          &17.60    &22.77       &62.09         &80.59         &74.39         &66.67        &31.67        &62.05        &56.74 \\ 
                TangentConv         &Block           & 52.6       & 62.2    & 90.5      &97.7       &74.0      &0.0           &20.7     &39.0        &31.3          &69.4          &77.5          &38.5         &57.3         &48.8         &39.8\\ 
                PointCNN            &Block           & 57.3       & 63.9    & 92.31     &98.24      &79.41     &0.00          &17.60    &22.77       &62.09         &80.59         &74.39         &66.67        &31.67        &62.05        &56.74\\ 
                PointWeb            &Block           & 60.3       & 66.6    & 91.95     &98.48      &79.39     &0.00          &21.11    &59.72       &34.81         &88.27         &76.33         &69.30        &46.89        &64.91        &52.46\\ 
            \midrule 
                baseline         &Block           & 61.8     & 70.3      & 91.6      &97.3       &81.6      &0.00          &14.4     &62.8        &58.4          &86.0          &77.4          &66.3         &56.9         &61.3         &49.1\\ 
                \bf{DGFA-Net}       &\bf{Block}    & \bf{65.8}  & \bf{73.8} &\bf{92.9}  &\bf{98.5}  &\bf{82.7} &0.00     &19.6&\bf{64.5}   &61.7     &87.4     &79.6     &68.9    &71.6    &\bf{75.1}    &53.5\\
            \bottomrule
        \end{tabular}
\end{table*}


\begin{table*}[htb]
    \footnotesize
    \caption{Detailed semantic segmentation results (\%) on ShapeNetPart. }
    \centering
    \begin{tabular}{m{2.2cm}@{}|@{}m{1.0cm}<{\centering}@{}|@{}m{1.0cm}<{\centering}@{}|@{}m{0.8cm}<{\centering}@{}
        @{}m{0.8cm}<{\centering}@{}@{}m{0.8cm}<{\centering}@{}@{}m{0.8cm}<{\centering}@{}
        @{}m{0.8cm}<{\centering}@{}@{}m{0.8cm}<{\centering}@{}@{}m{0.8cm}<{\centering}@{}
        @{}m{0.8cm}<{\centering}@{}@{}m{0.8cm}<{\centering}@{}@{}m{0.8cm}<{\centering}@{}
        @{}m{0.8cm}<{\centering}@{}@{}m{0.8cm}<{\centering}@{}@{}m{0.8cm}<{\centering}@{}
        @{}m{0.8cm}<{\centering}@{}@{}m{0.8cm}<{\centering}@{}@{}m{0.8cm}<{\centering}}
            \toprule
                Method&Class mIoU &Ins. mIoU & aero & bag & cap & car & chair & ear phone &guitar & knife & lamp &laptop &motor &mug&pistol&rocket&skate board &table\\
            \midrule
            PointNet             & 80.4      & 83.7 &83.4 &78.7 &82.5 &74.9 &89.6 &73.0 &91.5 &85.9 &80.8 &95.3 &65.2 &93.0 &81.2 &57.9 &72.8 &80.6\\ 
            PointNet++           & 81.9      & 85.1 &82.4 &79.0 &87.7 &77.3 &90.8 &71.8 &91.0 &85.9 &83.7 &95.3 &71.6 &94.1 &81.3 &58.7 &76.4 &82.6\\
            PCNN                 & 81.8      & 85.1 &82.4 &80.1 &85.5 &79.5 &90.8 &73.2 &91.3 &86.0 &85.0 &95.7 &73.2 &94.8 &83.3 &51.0 &75.0 &81.8\\
            DGCNN                & 82.3      & 85.1 &84.0 &83.4 &86.7 &77.8 &90.6 &74.7 &91.2 &87.5 &82.8 &95.7 &66.3 &94.9 &81.1 &63.5 &74.5 &82.6\\ 
            PointConv            & 82.8      & 85.7 &- &- &- &- &- &- &- &- &- &- &- &- &- &- &- &-\\ 
            InterpCNN            & 84.0      & 86.3 &- &- &- &- &- &- &- &- &- &- &- &- &- &- &- &-\\
            KPConv               & \bf{85.1} & \bf{86.4} &\bf{84.6} &\bf{86.3} &87.2 &\bf{81.1} &91.1 &\bf{77.8} &\bf{92.6} &88.4 &82.7 &\bf{96.2} &\bf{78.1} &\bf{95.8} &\bf{85.4} &\bf{69.0} &\bf{82.0} &\bf{83.6}\\
            \midrule  
            baseline          & 83.0      & 85.5      &83.9 &80.6 &88.2 &78.9 &90.8 &77.1 &91.0 &88.2 &84.0 &96.0 &71.6 &94.6 &82.7 &61.9 &76.4 &82.2\\
            \bf{DGFA-Net}        & 83.8      & 86.3      &84.3 &85.1 &\bf{88.3} &79.0 &\bf{91.6} &77.5 &91.9 &\bf{88.9} &\bf{84.8} &95.7 &72.5 &94.8 &83.7 &62.5 &77.0 &83.2\\
            \bottomrule
    \end{tabular}
\end{table*}
\begin{table*}[htb]
    \footnotesize
    \caption{Detailed semantic segmentation results (\%) on Toronto-3D.}
    \centering
    \begin{tabular}{m{2.8cm}@{}|c|c|cccccccc}
            \toprule
                Method&OA &mIoU & Road & Rd mrk. & Natural & Building & Util. line & Pole &Car & Fence \\
            \midrule
            PointNet++         &91.2 &56.6            &91.4 &7.6  &89.8      &74.0       &68.6      &59.5      &54.0     &7.5 \\ 
            PointNet++(MSG)    &90.6 &53.1            &90.7 &0.0  &86.7      &75.8       &56.2      &60.9      &44.5     &10.2\\ 
            DGCNN              &89.0 &49.6            &90.6 &0.4  &81.3      &64.0       &47.1      &56.9      &49.3     &7.3\\
            KPConv             &91.7 &60.3            &90.2 &0.0  &86.8      &86.8       &\bf{81.1} &\bf{73.1} &42.9     &\bf{21.6}\\ 
            MS-PCNN            &91.5 &58.0            &91.2 &3.5  &90.5      &77.3       &62.3      &68.5      &53.6     &17.1\\ 
            TGNet              &91.6 &58.3            &91.4 &10.6 &91.0      &76.9       &68.2      &66.3      &54.1     &8.2\\
            MS-TGNet           &91.7 &61.0            &90.9 &18.8 &\bf{92.2} &80.6       &69.4      &71.2      &51.1     &13.6\\
            \midrule
            baseline        &88.9 &57.1            &86.2 &50.4 &65.1      &\bf{92.0}  &49.0      &43.7      &68.3     &1.9 \\
            \bf{DGFA-Net}     &\bf{94.8} &\bf{64.3}   &\bf{96.0} &\bf{59.7}  &90.0       &80.2      &29.4      &62.0      &\bf{90.1} &6.8\\
            \bottomrule
    \end{tabular}
\end{table*}

\end{document}